\documentclass{article}

\usepackage{PRIMEarxiv}

\usepackage[utf8]{inputenc} 
\usepackage[T1]{fontenc}    
\usepackage{hyperref}       
\usepackage{url}            
\usepackage{booktabs}       
\usepackage{amsfonts}       
\usepackage{nicefrac}       
\usepackage{microtype}      
\usepackage{lipsum}
\usepackage{fancyhdr}       
\usepackage{graphicx}       
\graphicspath{{media/}}     
\usepackage{bibentry}
\usepackage{natbib}

\title{Stable Rivers: A Case Study in the Application of Text-to-Image Generative Models for Earth Sciences} 

\author{
  C. Kupferschmidt$^1$,
  A.D. Binns$^1$,
  K.L. Kupferschmidt$^{1,2}$, and
  G.W. Taylor$^{1,2}$\\
  $^1$School of Engineering, University of Guelph, Guelph, Ontario, Canada.\\
  $^2$Vector Institute for Artificial Intelligence, Toronto, Ontario, Canada.\\
  \texttt{\{kupfersc, abinns, kupfersk, gwtaylor\}@uoguelph.ca} 
  \\ 
  \\
  \textbf{Corresponding author:} Cody Kupferschmidt (kupfersc@uoguelph.ca)
}

\begin{document}
\maketitle

\begin{abstract}
Text-to-image (TTI) generative models can be used to generate photorealistic images from a given text-string input. These models offer great potential to mitigate challenges to the uptake of machine learning in the earth sciences. However, the rapid increase in their use has raised questions about fairness and biases, with most research to-date focusing on social and cultural areas rather than domain-specific considerations. We conducted a case study for the earth sciences, focusing on the field of fluvial geomorphology, where we evaluated subject-area specific biases in the training data and downstream model performance of Stable Diffusion (v1.5). In addition to perpetuating Western biases, we found that the training data over-represented scenic locations, such as famous rivers and waterfalls, and showed serious under- and over-representation of many morphological and environmental terms. Despite biased training data, we found that with careful prompting, the Stable Diffusion model was able to generate photorealistic synthetic river images reproducing many important environmental and morphological characteristics. Furthermore, conditional control techniques, such as the use of condition maps with ControlNet were effective for providing additional constraints on output images. Despite great potential for the use of TTI models in the earth sciences field, we advocate for caution in sensitive applications, and advocate for domain-specific reviews of training data and image generation biases to mitigate perpetuation of existing biases.

\end{abstract}


\section{Introduction}
The earth sciences field has seen a rapid uptake of machine learning (ML) technologies in recent years, in large part due to the increasing availability of “big data” from sources such as multi-spectral satellite imagery and remote sensing \citep{Yuan2020-kz}. However, difficulties in encoding domain knowledge \citep{Camps-Valls2020-pd}, a lack of benchmarking datasets \citep{Tuia2021-uf, Bergen2019-wb}, and predominantly unlabeled data \citep{Bergen2019-wb}, continue to limit uptake of ML models for many applications. Multimodal models, which allow for multiple data types (e.g. tabular data, text, image/raster data) and provide the potential for semantic interactions have been identified as a high-potential area for earth sciences research \citep{Tuia2021-uf}. Recent models such as RSVQA \citep{Lobry2020-oa}, which allows users to ask text-based questions about the semantic content of remote sensing imagery, have shown great promise in mitigating challenges around encoding domain knowledge. Advocacy by researchers for increased data openness and sharing \citep{Gil2016-xf, Gutierrez2021-ih}, and an increasing number of benchmarking datasets, are also helping to address data availability challenges.
However, for many specialized applications within the earth sciences field, data availability continues to be an issue. Generative models, which can be used to create synthetic data to augment limited existing datasets, have also been identified as a research area with high potential \citep{Bergen2019-wb}. Synthetic data is regularly used for many non-ML cases in the earth sciences \citep{Bunel2021-tm, Dawson2023-xr}, and recent research has demonstrated cases where the use of synthetic imagery in pre-training improves downstream model performance \citep{He2022-uu}.

Recent advances in text-to-image (TTI) models, which are both generative and multimodal, offer the potential to address multiple challenges currently faced in the earth sciences, since they allow users to interact with the models using natural language and can be used to generate synthetic photorealistic images. TTI generative models can be used to generate images based on a given conditional input, such as a caption-style text string known as a prompt. In particular, developments in diffusion-based approaches \citep{Yang2023-ei} that rely upon latent-space denoising autoencoders \citep{Rombach2022-dk} have allowed models to achieve photorealistic quality in image generation, and state-of-the performance \citep{Yasunaga2023-lb, Bao2022-jn} on common test datasets such as MS-COCO \citep{Lin2014-vi} and CUB \citep{Wah2011-rn}. Public awareness of TTI generative approaches is also on the rise, as evidenced by the recent popularity of commercially available models including DALL-E \citep{Ramesh2022-xn}, Imagen \citep{Saharia2022-ns}, and Stable Diffusion \citep{Rombach2022-dk}.

Although both the rise in popularity and performance of recent TTI models on benchmark datasets is impressive, questions remain about the limits to practical and safe use cases for the technology. Because ML models are typically trained by ingesting large amounts of training data and optimizing parameters to maximize specific objective functions, they are susceptible to perpetuating biases and inequalities that are present in the data or objective functions. For example, several popular TTI diffusion models have been shown to lack diversity and be biased towards over-representing concepts such as whiteness and masculinity across targets \citep{Luccioni2023-ez}. Despite these known limitations, major graphic design and digital media companies such as Adobe \citep{Adobe2023-a}, Canva \citep{Canva2022-qn}, and Shutterstock \citep{Shutterstock2023-dm} already offer tools that allow users to generate AI images on their platforms, with Adobe \citep{Adobe2023-b} reporting that more than one billion images had been generated as of July 12, 2023. The rapid use and increasingly widespread deployment of these technologies raises questions surrounding potential amplification of existing biases and inequalities \citep{Luccioni2023-ez}.

To-date, much focus on the implications of ML models, and more specifically TTI models, has focused on social and cultural biases \citep{Luccioni2023-ez, Shankar2017-yf, DeVries2019-oy}. As the domains of use for generative multimodal models expand, researchers in the earth sciences field have also raised concerns about the need to evaluate domain-specific biases \citep{Tuia2021-uf, Bergen2019-wb}. The advent of TTI models presents an exciting opportunity to make ML models more accessible for scientists, since the main interaction between the user and model can be performed using text-based prompt rather than a programming language. However, increasing uses of these models by non-computer scientists, who do not fully understand the training process, will come with an increased risk of perpetuating existing biases within the models.

The field of fluvial geomorphology is a subset of of the earth sciences field that is focused on the study of the geometry and behaviour of rivers. Several studies \citep{Dupont2018-ic, Chan2017-ok} in the field of generative modelling have focused on replicating the geomorphological patterns associated with river meandering using generative adversarial networks. Similar to many other earth science areas, the field of fluvial geomorphology continues to rely heavily on photographic images for many purposes, such as monitoring landscape evolution \citep{Cerney2010-lu} or assisting in the classification of river morphologies \citep{Rosgen1996-if}. Use cases for data augmentation using TTI models to produce synthetic image datasets for other earth sciences applications are already emerging \citep{nathanail2023geo}, and numerous use cases exist for fluvial geomorphology such as generating images for natural channel design, river restoration, and morphological evolution.

To the best of the authors’ knowledge, no studies to-date have used diffusion-based TTI models to generate images of rivers with specific morphological properties. Additionally, no assessment of the biases and limitations of any existing TTI models specific to fluvial geomorphology have been conducted. In the present study we explore a subset of the training data used for the Stable Diffusion model \citep{Rombach2022-dk} to identify differences in their representations for applications specific to the field of fluvial geomorphology. Biases and their downstream impacts on image generation associated with the inclusion of general, morphological, and environmental terms in text prompts are also evaluated for Stable Diffusion (v1.5) \citep{Rombach2022-dk}. Finally, we evaluate if ControlNet Canny edge conditioning can be used with the evaluated model to generate photograph-like images of rivers that contain key morphological and environmental characteristics relevant to the field of fluvial geomorphology.

\section{Background}
\subsection{Fluvial Geomorphology}
The field of fluvial geomorphology is based on the study of interactions between sediment and water in rivers over time. In particular, fluvial geomorphology focuses on the study of rivers with erodible beds and banks, known as alluvial rivers. Within these rivers, relationships exist between the sediment transport capacity of the flow and the supply of sediment. The term “stable” is often used to refer to sections of rivers that may experience temporary localized erosion or deposition of sediment, but where no net scour or aggradation is observed over longer periods of many years or decades. In contrast, when the sediment transport capacity of a river is not matched with the sediment supply, the term “unstable” is often used, which describes a river that may experience trends in morphological adjustment over time due to sediment deposition or erosion, potentially resulting in slope changes due to aggradation or degradation, channel widening or narrowing, and/or length adjustment until a new equilibrium is established \citep{Mackin1948-it}. 

To monitor changes in alluvial rivers over time, as well as more broad measures of watershed health, many protocols for assessing the stability and morphology of streams have been established. In particular, “rapid” or “visual” geomorphic assessment guidelines have emerged as a key tool for conducting preliminary assessments of stream stability and health, with various states, provinces, and government agencies each producing their own guides (e.g. \citep{Ontario_Ministry_of_Natural_Resources2010-fk},\citep{Natural_Resources_Conservation_Service1998-tz}, \citep{Maine_Department_of_Environmental_Protection2010-oz}, and \citep{Vermont_Agency_of_Natural_Resources2009-qu}).   

Many of these assessment techniques rely heavily on documentation and monitoring of active or ongoing fluvial processes through photographs. Therefore, an understanding of the ability of TTI models to generate images that accurately produce key geomorphic characteristics of fluvial river systems is essential for incorporating these models into applications for fluvial geomorphology.

\subsection{Stable Diffusion and ControlNet }
Stable Diffusion is a popular and recently-developed diffusion-based architecture that relies upon sequential application of denoising autoencoders to perform high resolution image synthesis \citep{Rombach2022-dk}. Rather than operating in the high-dimensional pixel space, the model uses lower-dimensional latent space autoencoders to balance computational requirements with image quality. Stable Diffusion is able to accept several conditioning mechanisms to control the image synthesis process, including text, which allows it to function as a TTI model. When used as a TTI model, Stable Diffusion extracts semantic latent-space embeddings from text prompts and generates a novel image by decoding these embeddings to a user-interpretable pixel space. 

Weights for trained versions of Stable Diffusion that accept text prompt conditioning have been made publicly available online \citep{Rombach2022-dk}. The publicly available model weights for Stable Diffusion (v1.1) were trained on a subset of the LAION-5B image dataset \citep{Schumann}, using LAION-2B-en, which contains 2.32 billion images with corresponding English text captions. A more recent version of the model (v1.5), which was used in the present study, includes additional training resumed from the v1.1 checkpoint using the LAION-Improved-Aesthetics (v1.2) and LAION-Aesthetics-V2-5+ (v1.5) datasets which are derived from further filtering of the original dataset to only include images larger than 512x512 with high estimated aesthetic scores (> 5.0) and low estimated watermark probabilities (< 0.5). 

ControlNet is a neural network structure that allows for additional control of image generation for pretrained diffusion models \citep{Zhang2023-kb} using conditional input maps, e.g. Canny edge maps or human sketches, hereafter termed “condition maps” for short. The structure can be used to guide the output of large pre-trained models such as Stable Diffusion, without the requirements of re-training the model for specific tasks. This is achieved by creating a trainable copy of weights from the original model that can be trained on condition maps and applying these weights to selected blocks of the original model. This technique allows task-specific conditioning, even on small (<50k) datasets. A number of pre-trained ControlNet models that can be used to augment Stable Diffusion are available online for condition maps such as edge maps, depth maps, and normal maps. 

\subsection{Biases and Inequalities }
The data-centric nature of ML algorithms makes them specifically sensitive to biases and inequalities. ML models typically ingest large amounts of training data, and learning algorithms optimize their parameters to achieve a specific practitioner-defined objective.  Biases can be introduced at a number of stages throughout this process including data collection, data preparation, model development, model evaluation, model post-processing, and model deployment \citep{Suresh2019-jb}. Each potential source of bias may result in ML models producing unfair or inequitable results. 

The concept of biases being perpetuated from training and evaluation datasets are some of the most widely discussed biases in ML. These biases typically arise during the data collection and preparation phase, and include historical, representation, and measurement biases \citep{Suresh2019-jb}. Historical biases are caused by unequal real-world distributions across a variable, while representation biases are caused by unequal sampling. Measurement biases arise during the feature or label generation process. Since many of the training datasets used for ML models are human-curated, some of the biases in training data are likely to be introduced by human decisions. These inequalities can become further exaggerated when decisions made by ML algorithms influence future human behavior, biasing future data generating processes. 

There are several prominent examples of widely used benchmarking datasets that demonstrate inequalities in data representation, such as ImageNet and Open Images, both of which demonstrate a strong bias towards Western cultures \citep{Shankar2017-yf, DeVries2019-oy}. The LAION-400M dataset, a precursor to the LAION-2B-en dataset, which was used to train the Stable Diffusion TTI model \citep{Rombach2022-dk}, has been found to contain troublesome imagery and caption pairs with misogynistic, pornographic, and malignant stereotypes \citep{Birhane2021-ve}. Many of these biases present in the training data can be carried through into model outputs. For example, the Stable Diffusion \citep{Rombach2022-dk}model card acknowledges that default outputs are biased towards western and white cultures, while overrepresentation of whiteness and masculinity have been identified in other popular generative TTI models \citep{Luccioni2023-ez}. 

However, there are many biases that exist beyond the more commonly studied social and cultural biases. Within the environmental science field, opportunistic sampling may lead to the collection of data without proper consideration for spatial and/or geographic representation. Sampling biases have been observed due to geographic factors, such as proximity to roads, and environmental factors such as weather (precipitation and temperature), with significant impacts to online datasets \citep{Cosentino2021-bp}. Similar biases known as the “cottage effect” have also been observed in citizen science programs that rely on opportunistic or incidental sampling procedures \citep{Millar2019-bd}, where lakes with attractive characteristics, proximal to heavily populated areas, and leisure destinations are over-represented. 

Within the field of fluvial geomorphology, many systemic biases are known to exist. For morphological factors, studies on the perceptions of beauty have found that braided river channels containing gravel bars are seen as less aesthetic than those without \citep{Le_Lay2013-ui}, while other research has demonstrated that freshwater scientists hold biases towards single-thread river channels with symmetrical meander planforms for river restoration projects \citep{Wilson2020-zk}. Some research even suggests that biases in channel restoration projects may be tied back to 18th Century English landscape theories, which treated serpentine channels as ideal \citep{Podolak2016-ju}. Human perceptions of environmental factors such as streamflow are also subject to biases, with one study finding that the most aesthetic flows for a river located in northern Colorado typically occurred twice a year, and were associated with the rising and fall limbs of the spring runoff hydrograph \citep{Brown1991-gw}. 

\section{Methodology}
The current study involved three main research activities: 

\begin{enumerate}
  \item \textbf{Training Data Representation:} Assess the representation of images with captions containing various terminology relevant to the field of fluvial geomorphology in the Stable Diffusion training data.

  \item \textbf{Biases and Model Impacts:} Evaluate biases and impacts associated with the use of subject-area specific terminology on images generated by Stable Diffusion.

  \item \textbf{Image Reconstruction:} Assess if the Stable Diffusion model can be used with prompts and additional input conditional control (ControlNet) to reconstruct images of rivers.
\end{enumerate}

Commonly used descriptors in the field of geomorphology were used across all three research activities. To simplify the assessment, we grouped descriptors into the following three categories: 
\begin{itemize}
  \item \textbf{General:} watercourse descriptor, location name (country, state/province, famous river).

  \item \textbf{Morphological:} landscape, channel gradient, channel dimension, and geology.

  \item \textbf{Environmental:} water color, flow condition, weather, season.
\end{itemize}

\subsection{Training Data Representation}

In order to evaluate the relative weights of the selected descriptors in the training data used for Stable Diffusion, we used the Improved-Aesthetics-6+ Datasette browser from Willison and Baio (2022). This browser contains data for 12 million images from LAION-Aesthetics-6+, a subset of LAION-5B \citep{Schumann}, which contains approximately 0.5\% of the data used in the initial training of Stable Diffusion (v1-1) and 2\% of the data used in each of the subsequent checkpoints of more recent releases (v1.2, v1.5). Although it makes up a small subset of the total training data used in the three most recent checkpoints, the dataset was assumed to be generally representative of the training dataset with a slight bias towards more aesthetic images. The aesthetic score of an image is a subjective rating in which users are asked, “How much do you like this image on a scale from 1 to 10?”, and a model to predict this score was used in the filtering process for the creation of the LAION-Aesthetics datasets. 

For each of the descriptor terms, we conducted a search to return the number of rows in the dataset with a matching caption. For categories other than watercourse descriptor, the word “river” was added to the search term, and a subsequent search was conducted.  

Within the general category of descriptors, we performed additional analyses for location names. For country names we selected several countries in each continent that are well-known for major and/or scenic rivers, as we expected strong biases to be associated with each of these rivers. For each continent the most populous country was also included. The number of training examples for each country were compared with both population \citep{World_Bank2022-pb} and land area \citep{World_Bank2020-gl} data. To provide a more regional assessment, the names of selected provinces and states within Canada and the USA were also evaluated. The most populous province and states were evaluated, along with provinces and states well-known for their scenic rivers. The names of several of the most famous rivers from each continent were also evaluated.

\subsection{Biases and Model Impacts}
Biases inherent to Stable Diffusion (v1.5) were evaluated through prompt variation. Each of the descriptor terms from the Training Data Representation task was used as a prompt, with \textit{n=30} images generated for each prompt. For terms in categories other than watercourses, the term “river” was added to the prompt text. Consistent seeding was used during prompt variation to allow for comparison between images. To increase the probability that the model output photorealistic images of rivers, the following negative prompts were also used for all images generated: grayscale, low quality, painting, people. 

A qualitative visual review of image changes due to prompt variation was performed, with a focus on morphological and environmental features of the river images. A quantitative analysis of changes in semantic image characteristics was performed using the BLIP model \citep{Li2022-rr} to perform image captioning, and to extract the most common caption words for each unique prompt. The following common words were excluded from the list: the, in, it, by, at, is, and, a, of, on, with, that. We also computed the CLIP \citep{Radford2021-ko} score which uses the cosine distance between feature embeddings between the images generated for each prompt and compared them with images generated for the same seed for the baseline or default prompt of “river”. This serves as a numerical metric to quantify the downstream impact of the prompt variation on the image semantics. Higher scores indicate greater image similarity, with a maximum score of 1.0 indicating semantically identical images.

\subsection{Image Reconstruction}
Based on inequalities in training data and model biases identified in the first two tasks, we evaluated the impact of using additional conditional controls to mitigate or reduce biases in generated images. Specifically, we wanted to see if it was possible to reproduce images using TTI models based only on a simplified representation of the image. 

A set of \textit{n=40} images of natural rivers collected by the authors that have not been seen by the Stable Diffusion or ControlNet models during the training process were used for this task. Images were resized to 512 pixels in their shortest direction, and Canny edges \citep{Canny1986-fy}, representing visible edges from with the natural river images, were extracted using the default function included with ControlNet (v1.0). 

For each of the natural river images, we used ControlNet (v1.0) with Stable Diffusion (v1.5) to produce images using both the Canny edges and a text prompt. Three text prompts were evaluated: a user specified prompt that generated the best reproduction of the original image determined through trial and error, a caption from the original image generated using BLIP \citep{Li2022-rr}, and a default prompt of “river”. For the model we used 20 inference steps, with consistent seeding during prompt variation to allow for comparison between images. The same negative prompts were used as with the previous task: grayscale, low quality, painting, people. For each prompt and Canny edge pair, we generated \textit{n=5} images. 

The output images generated by the model were compared with the original images to determine the degree to which the original image was reproduced. In this approach, the original image serves as an “oracle”, representing a theoretical upper bound on performance. This was done both qualitatively as well as quantitatively by computing the CLIP \citep{Radford2021-ko} score between the original and generated images. 

The current approaches use Canny edge condition maps as a proof of concept for studying the conditional control of TTI generative models for fluvial geomorphology. However, the use of conditional controls such as ControlNet are general approaches that can accept many different types of conditioning \citep{Zhang2023-kb}. In future work, other models could be trained to accept condition maps more relevant to the field of fluvial geomorphology, including digital elevation models, land use maps, soil types, or vegetation. 

\section{Results}

\subsection{Training Data Representation}
The number of rows returned for each of the descriptor terms in the LAION-Aesthetics-6+ dataset are shown in Table \ref{fig:table_1} through Table \ref{fig:table_3}.

\begin{table}
 \centering
  \includegraphics[width=0.8\textwidth]{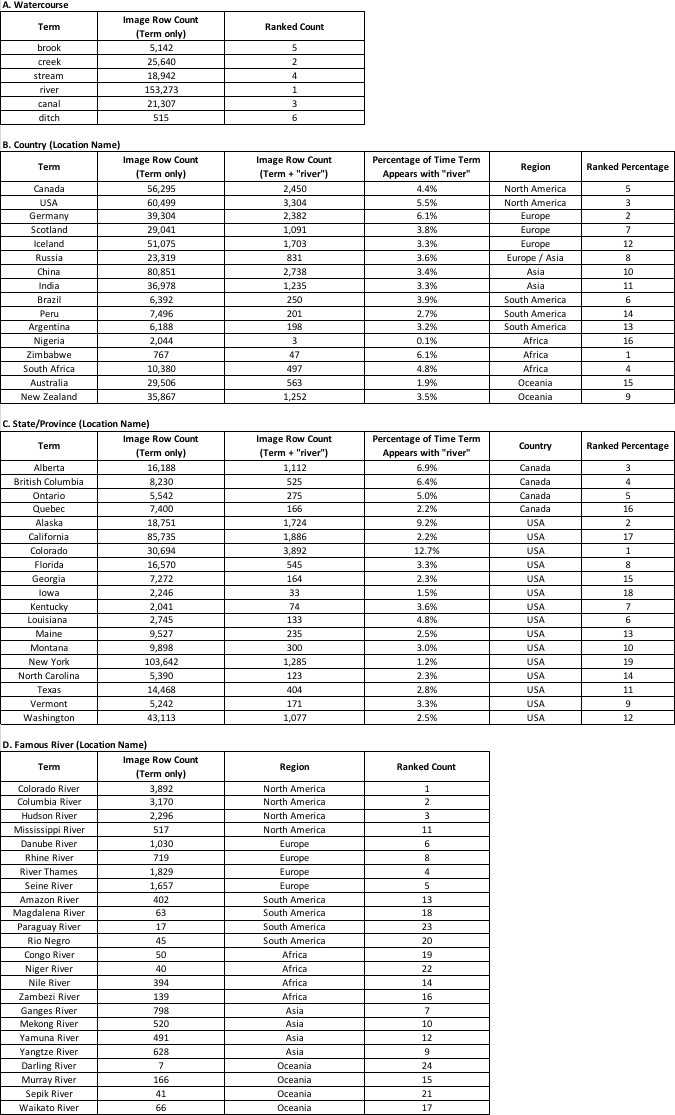} 
  \caption{Number of training images for Stable Diffusion included in the LAION-Aesthetics-6+ database by search term for A. watercourse, B. country, C. state/province, and D. famous river.}
  \label{fig:table_1} 
\end{table}


For the general descriptors (Table \ref{fig:table_1}), the term “river” was the most commonly used to refer to natural watercourses (\(\sim10^5\) results) in the training data, with “creek” and “stream” having a representation in the data of approximately one order of magnitude lower (\(\sim10^4\) results). The term “brook” not only had lower representation in the training data (\(\sim10^3\) results), but also suffered from ambiguous results associated with the common human given name Brook. For human-made watercourses, the term “canal” returned the greatest number of results (\(\sim10^4\) results), approximately two orders of magnitude larger than “ditch” (\(\sim10^2\) results).

\begin{figure}
  \centering
  \includegraphics[width=0.8\textwidth]{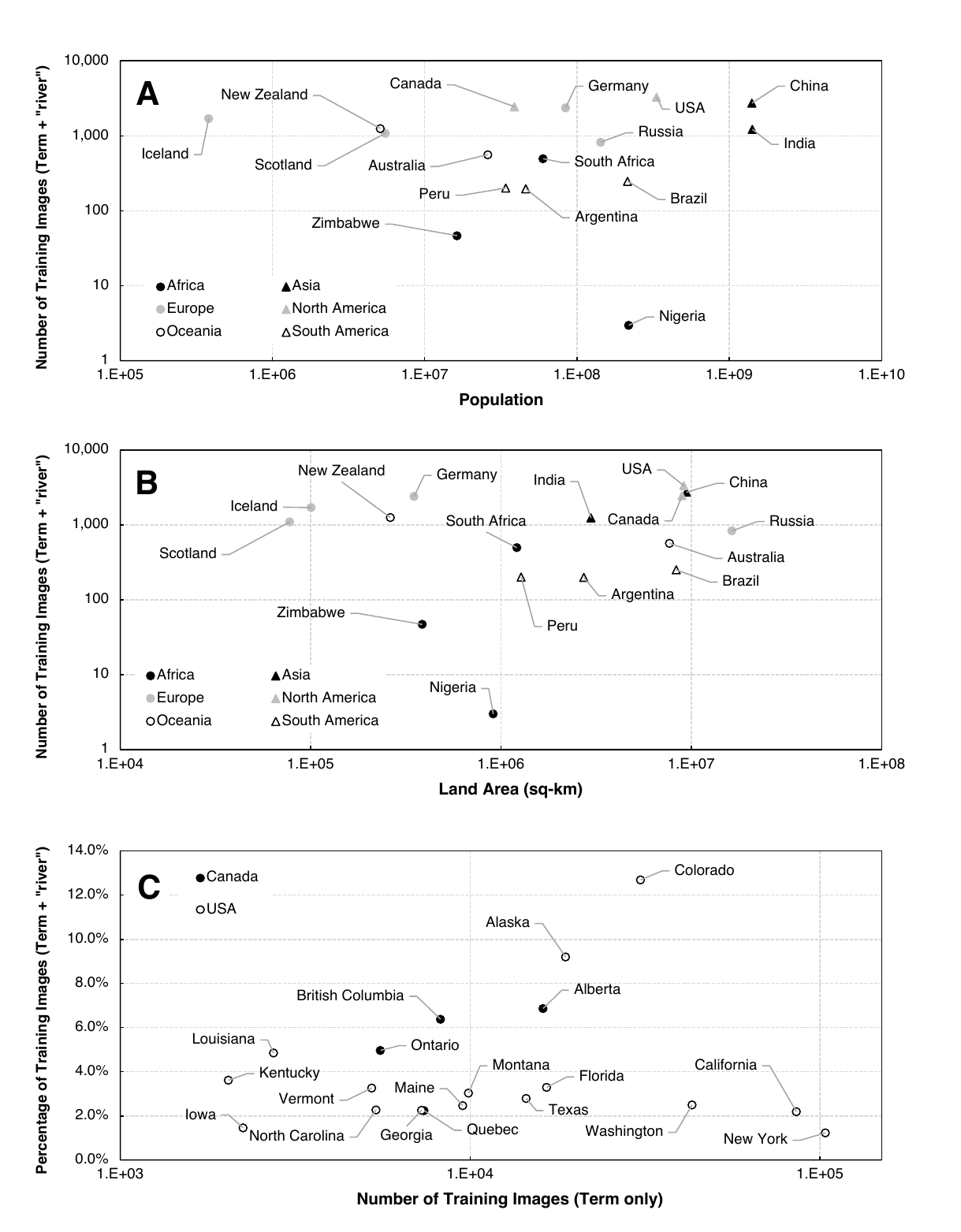} 
  \caption{Relationships between the number of training images by A. population, and B. land area for each evaluated country. C. Proportion of training images containing the term “river” for each evaluated province and state. Population and land area from World Bank Open Data (2022).}
  \label{fig-1} 
\end{figure}

We identified that the western biases known to be associated with Stable Diffusion (Rombach et al. 2022) are also present in river-specific images. Over-representation of rivers from North American and European countries in terms of absolute representation (number of training images returned in row count for term + “river”), representation by population (Figure \ref{fig-1}A), and representation by land mass (Figure \ref{fig-1}B) were identified in the training data, although China, and to a lesser extent India, also had moderate representation. On both a national (e.g. Iceland, New Zealand, Canada, Figure \ref{fig-1}B) and state/province level (e.g. Colorado, Alaska, Alberta, British Columbia, Figure \ref{fig-1}C), popular tourist destinations with well-known scenic rivers were observed to be over-represented in the training data. Results for famous river names also perpetuated known western biases, with results for data from North America and European rivers (\(\sim10^3\) results on average) approximately one order of magnitude higher than from the next highest region in Asia (\(\sim10^2\) results on average). Results from South America, Africa, and Oceania showed severe under-representation (\(\sim10^{1}\) to $10^2$ results each).

For morphological descriptors (Table \ref{fig:table_2}), we observed that the terms “forest,” “mountain,” “canyon,” “valley,” and “city” had the highest number of results in the landscape descriptor category of river images (\(\sim 10^3\) results). In the channel gradient category, the term “waterfall” had the highest representation (\(\sim 10^3\) results), with most other terms returning results one order of magnitude lower (\(\sim 10^2\) results). A notable outlier for channel gradients was the term “riffle,” which, despite being widely used in the field of fluvial geomorphology, is not common in general vernacular and had a representation several orders of magnitude lower (\(\sim 10^2\) results). All terms evaluated for channel dimensions returned similar numbers of results (\(\sim 10^2\) to \(\sim 10^3\) results). For geology descriptors, the term “rock” had the greatest representation for river images (\(\sim 10^3\) results), with the terms “sand,” “boulders,” “limestone,” and “granite” returning results about one order of magnitude lower (\(\sim 10^2\) results).

\begin{table}
 \centering
  \includegraphics[width=0.8\textwidth]{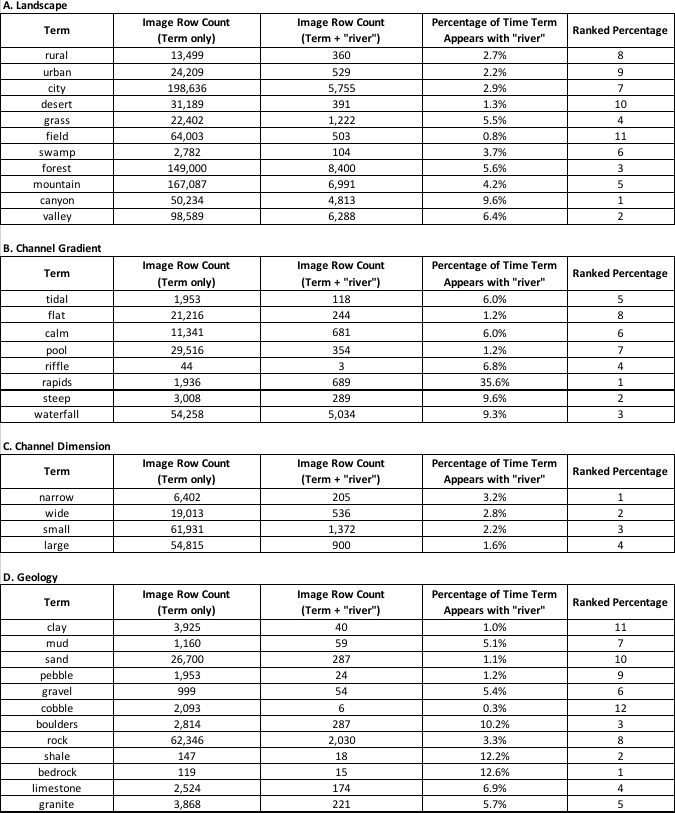} 
  \caption{Number of training images for Stable Diffusion included in the LAION-Aesthetics-6+ database by search term for A. landscape, B. channel gradient, C. channel dimension, and D. geology.}
  \label{fig:table_2} 
\end{table}

For the environmental descriptors (Table \ref{fig:table_3}), in the water color category, the terms “brown” and “blue” had the highest representation (\(\sim 10^3\) results), followed by “clear” and “whitewater” (\(\sim 10^2\) results). In terms of flow condition, the greatest number of river images were returned for the term “flowing” (\(\sim 10^3\) results), followed by “flooding” and “dry” (\(\sim 10^2\) results). For each of the weather categories (\(\sim 10^2\) results) and season categories (\(\sim 10^3\) results), a similar number of results were returned for each term within the respective category.

In general, of the evaluated river descriptor categories, those with the greatest representation in the training data included: watercourse descriptors, country names, famous river names (North America only), seasons, and landscapes.

\begin{table}
 \centering
  \includegraphics[width=0.8\textwidth]{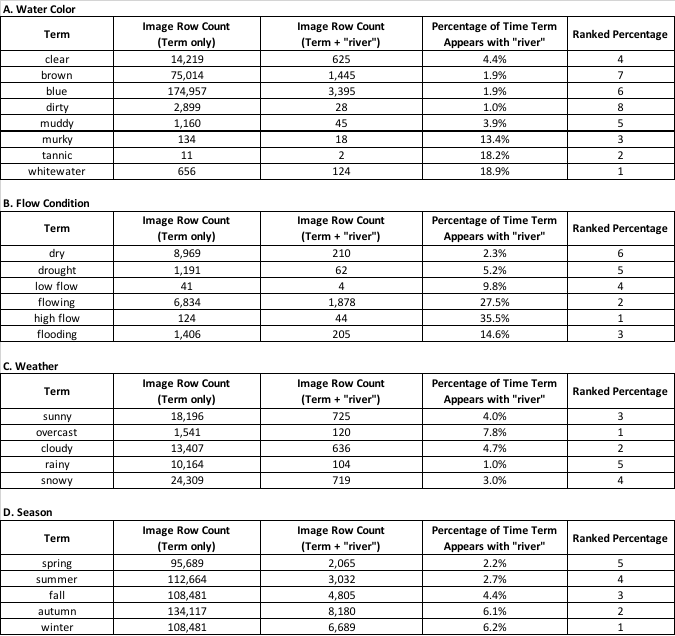} 
  \caption{Number of training images for Stable Diffusion included in the LAION-Aesthetics-6+ database by search term for A. water color, B. flow condition, C. weather, and D. season.}
  \label{fig:table_3} 
\end{table}

\subsection{Biases and Model Impacts}
\begin{figure}
  \centering
  \includegraphics[width=0.8\textwidth]{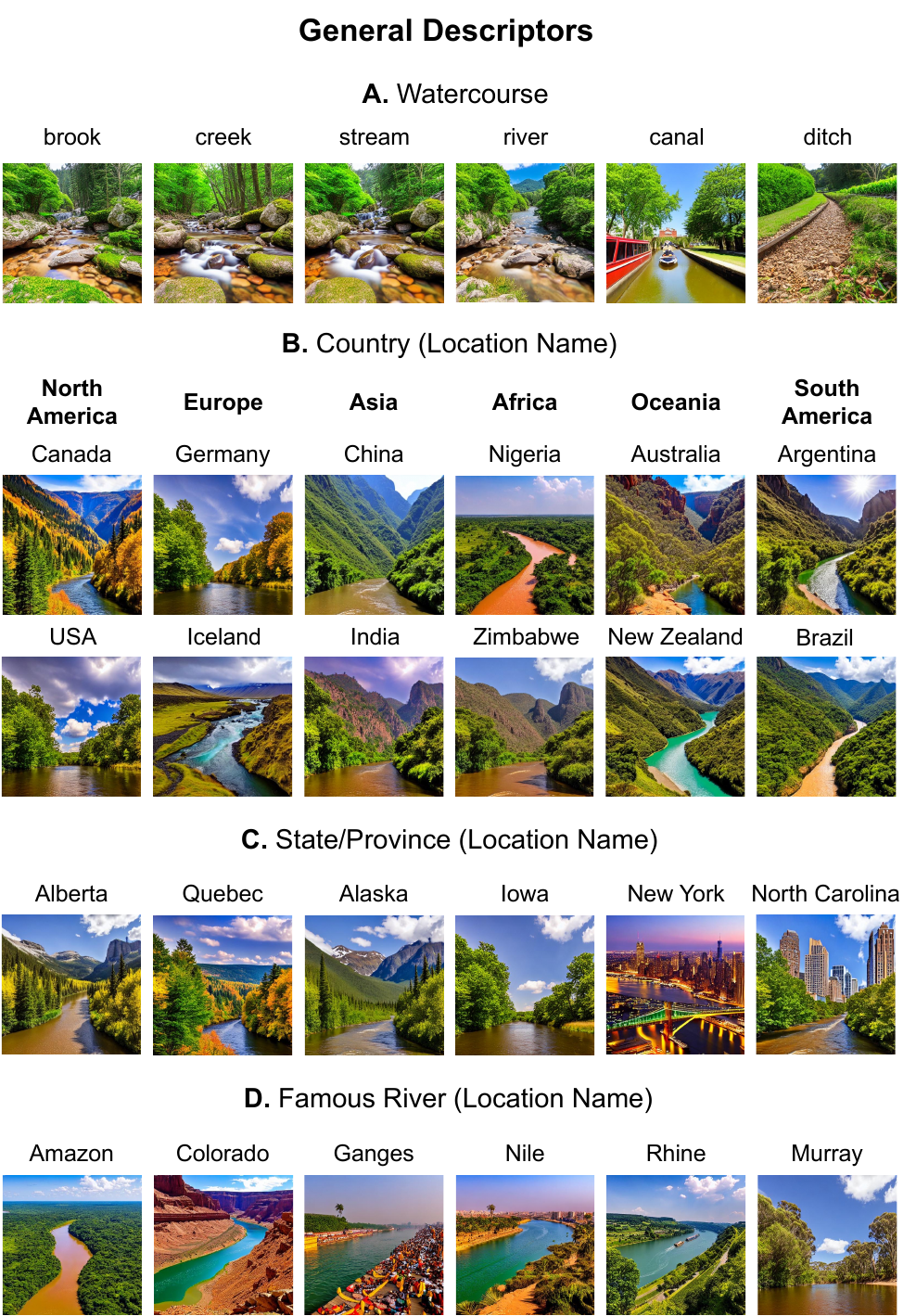} 
  \caption{Examples of the impact of prompt variation using general descriptors for unconditional image generation using Stable Diffusion (v1.5). Results are shown for terms from the categories of A. watercourse, B. country (name), C. state/province (name), and D. famous river (name)}
  \label{fig-2} 
\end{figure}

\begin{figure}
  \centering
  \includegraphics[width=0.8\textwidth]{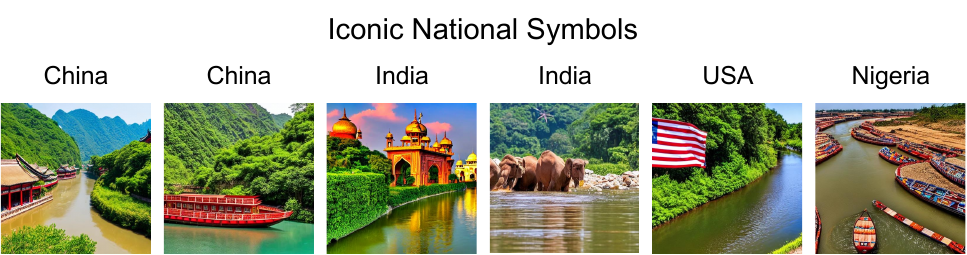} 
  \caption{Examples of the inclusion of iconic national symbols (e.g. traditional architecture, boats, local wildlife, and national flags) in generated river images that include country name in the prompt for unconditional image generation with Stable Diffusion (v1.5).}
  \label{fig-3} 
\end{figure}

\begin{figure}
  \centering
  \includegraphics[width=0.8\textwidth]{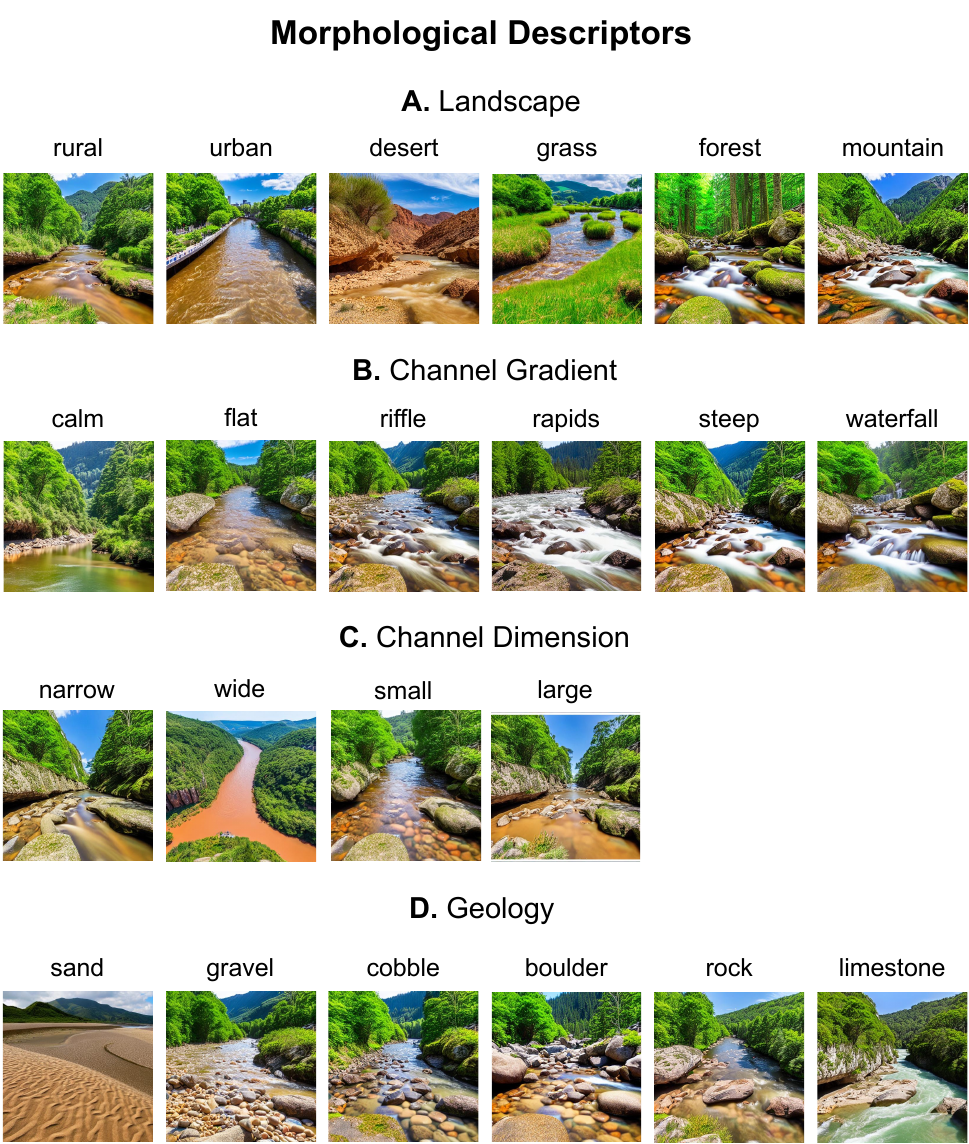} 
  \caption{Examples of the impact of prompt variation by adding each morphological descriptor to the term “river” for unconditional image generation using Stable Diffusion (v1.5). Results are shown for terms from the categories of A. landscape, B. channel gradient, C. channel dimension, and D. geology.}
  \label{fig-4} 
\end{figure}

\begin{figure}
  \centering
  \includegraphics[width=0.8\textwidth]{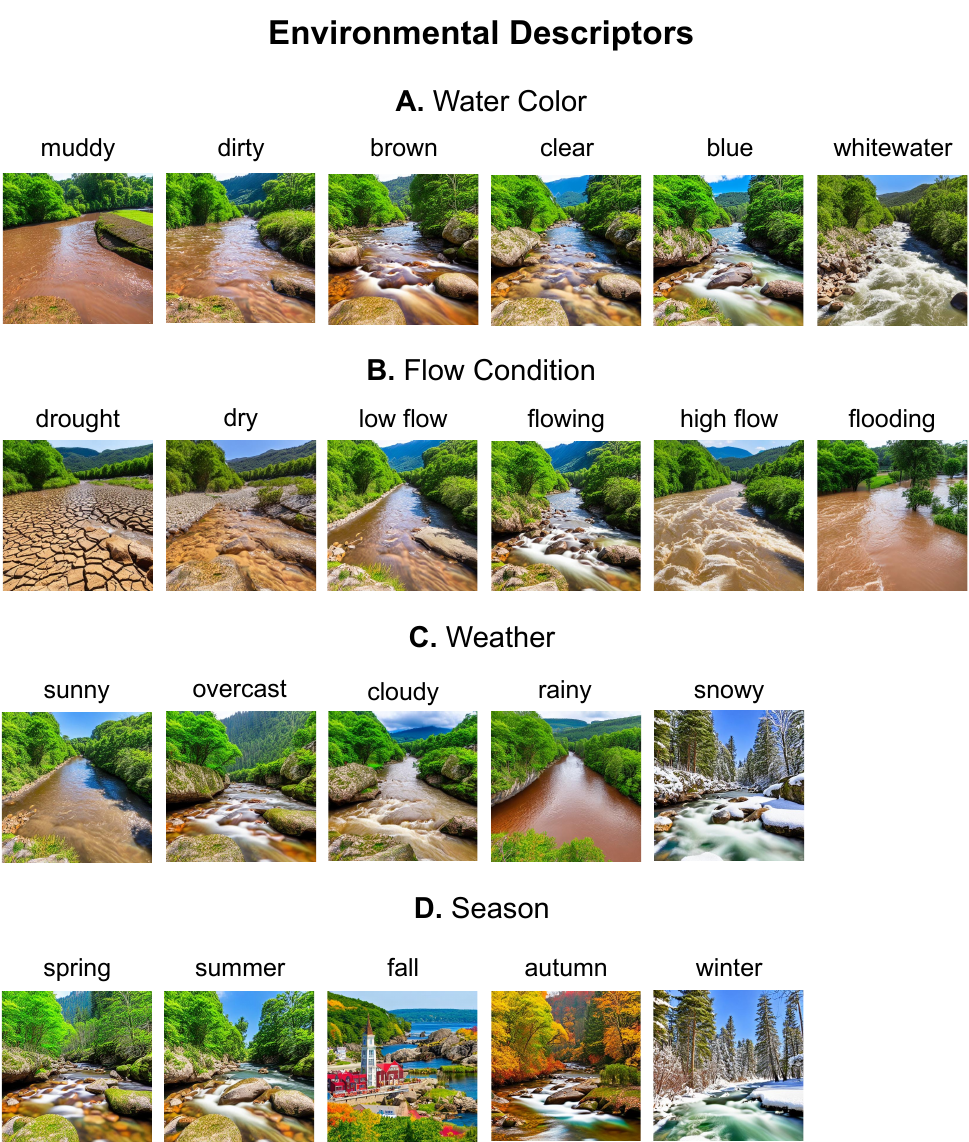} 
  \caption{Examples of the impact of prompt variation by adding each environmental descriptor to the term “river” for unconditional image generation using Stable Diffusion (v1.5). Results are shown for terms from the categories of A. water color, B. flow condition, C. weather, and D. season. }
  \label{fig-5} 
\end{figure}

Using the Stable Diffusion model, we generated \textit{n=30} images for each descriptor term. A subset of the resultant images for selected terms within each of the categories that demonstrate key findings are shown in Figure \ref{fig-2} through Figure \ref{fig-5}. 

For watercourse terms within the general descriptor category, prompt variation appeared to be an effective technique for capturing key differences typically associated with the various terms. Prompts using the terms “brook”, “creek”, and “stream”, often used to describe smaller tributaries, were observed to be more likely to produce images with small channels and steeper gradients when compared with the prompt “river” (Figure \ref{fig-2}A). The top-five BLIP caption words associated with the smaller channel terms included “stream”, “forest”, and “woods”, while prompts “stream” and “creek” generated images that were also associated with the word “small”.  Use of the term “canal” tended to produce images of watercourses with straightened channels and hardened or concrete-lined banks, which are often in urbanized landscapes (Figure \ref{fig-2}A). This urban bias was also present in the BLIP captions for “canal”, which included the words “city” and “park”. Use of the prompt “ditch” resulted in a high failure rate, and tended to produce images of footpaths rather than watercourses. The top BLIP caption associated with this prompt was “path”. 

On a country level, mountain landscapes were the most frequently generated type of landscape (e.g. Figure \ref{fig-2}B), even for countries with diverse geographies. Qualitatively, many image features produced by the model, such as the landscapes (e.g. few trees in Iceland, deciduous trees in Germany) and water colors (e.g. muddy rivers in Nigeria and Brazil, blue rivers in Iceland and New Zealand) were plausible for the target countries. 

However, in some cases the model perpetuates existing stereotypes by inserting iconic national symbols on river images in an attempt to make them better match the prompt description, such as the inclusion of traditional architecture, local wildlife, boats, or even national flags (Figure \ref{fig-3}). 

Similar to the country prompts, for the state/province place name descriptors the generated landscapes were generally plausible for the prompt state or province, with BLIP captions often containing the word “mountain” for mountainous regions, and “forest” or “woods” for non-mountainous regions. North Carolina and New York had a large number of images associated with urban environments, which was also seen in the BLIP captions containing the words “city”, “skyline”, and “downtown”. Regions well-known for outdoor recreation in mountainous areas had the highest CLIP scores (Maine: 0.94, Vermont: 0.92, Alberta: 0.91, and Colorado: 0.91) when compared with the default “river” image. Each of these regions contained the word “mountain” and aside from Colorado, also had the words “forest” or “woods” in their top-five BLIP terms, which is the same as the BLIP captions for the default “river” image. This suggests a strong similarity between the model interpretation of locations and the default interpretation of the term “river” in the Stable Diffusion model. 

For famous rivers, the model performed well, and was able to generate morphologically realistic images for most of the prompts (Figure \ref{fig-2}D). Many river images yielded BLIP captions that included the river name (e.g. Colorado, Nile, Rhine, Amazon), however, a number of rivers in Africa also included the word “amazon” in their BLIP captions. The CLIP scores associated with using specific river names as prompts were generally lower than country or state/province names, suggesting a greater impact on the output image. 
 
The Stable Diffusion model was generally effective in interpreting the morphological descriptors and was able to generate images that included many characteristics commonly associated with the terms. For the landscape descriptors (Figure \ref{fig-4}A), the terms “urban” and “city” result in the model generating river images with straightened channels, buildings, and other evidence of human land use, while other terms showed more natural planforms and landscapes. The model adjusted riparian zone vegetation, channel gradient, and geology in the generated images to meet the provided landscape descriptor. For certain terms such as “grass”, “field” and “desert” there was a relatively high failure rate, with more than 10\% of images generated by the model not containing an obvious river. The terms “mountain” and “rural” had the highest CLIP scores (0.93 and 0.92, respectively), indicating the most image similarity to those generated using the default prompt “river”. Despite a high failure rate for the prompt “sand”, the BLIP assessment showed that many generated images had a strong association with the word “wadi”, an Arabic term for river valley that is sometimes used to describe ephemeral watercourses. 

Through prompt variation involving channel gradient descriptors (Figure \ref{fig-4}B), it was possible to generate a series of images of rivers with gradually increasing gradients. The terms “calm” and “flat” were effective for generating river channels with low gradients, with “riffle”, “rapids” and “steep” resulting in moderate gradients, and “waterfall” producing the highest gradients. The images showed some inconsistencies in channel gradient for images generated using the term “riffle”, which was noted to have one of the lowest representations of all evaluated terms in the training data. The term “pool” is widely used in fluvial geomorphology to describe deep, slow moving sections within a river; however, when this term was included, the model exclusively generated images of swimming pools, and had a CLIP score of 0.62, indicating very dissimilar images to the default prompt. The BLIP captions showed that images generated using steeper gradient terms in the prompts had a stronger association with the word “mountains” than lower gradient terms. 

For the evaluated channel dimension descriptors (Figure \ref{fig-4}C), the terms “narrow” and “small” generally produced smaller channels, while “wide” produced larger channels. In the BLIP caption analysis, the term “small” was associated with the words “river” and “stream”, while the other channel dimension descriptors showed a stronger relationship to the word “river”. The use of the term “large” produced inconsistent results in terms of channel dimensions, and demonstrated a high failure rate by commonly producing non-river images, including some outputs that returned black images after being flagged by the model as not safe for work (NSFW). 

Geology descriptors were generally well-interpreted by the model, as seen in Figure \ref{fig-4}D, with gradually increasing bed material size from left to right. A high failure rate in generated images was observed for the term “sand”, with many of the generated images containing no river and/or text similar to “sand river”. CLIP scores were the highest for the terms “rock”, “shale”, and “cobble”, suggesting the greatest similarity to the default prompt. Geological descriptors were generally not present in the BLIP captions for the generated images.  

For the environmental descriptors, the Stable Diffusion model was generally able to generate images with water color corresponding to the input prompt descriptor (Figure \ref{fig-5}A). Increasing water clarity was often accompanied by increasing channel gradient, the presence of cobble or boulder channel geology, and more mountainous landscapes. This is also seen in the BLIP captions, with clearer water more strongly associated with “mountains” and dirtier water more strongly associated with “forest”, “trees”, or “woods”. The CLIP score for images generated using the descriptor “whitewater” (0.95) was the second highest of all evaluated prompts, suggesting a strong association between the term and the model’s default interpretation of a river. 

Stable Diffusion was also generally able to reproduce flow conditions (Figure \ref{fig-5}B), although differences between “low flow”, “flowing”, and “high flow” were not always consistent. From the BLIP captions, the term “drought” was associated with the words “mountain” and “desert”, while moderate to high flows were associated with “mountains” and “forest”. This is an important bias to note in the model results since drought is a relative term and may occur in any environment. Of all of the evaluated flow conditions, the term “flooding” had the lowest CLIP score.  

Although some weather conditions were reproduced by the model (Figure \ref{fig-5}C), none of the images generated for the term “rainy” showed precipitation occurring. Additionally, while other descriptors such as water color and flow condition had relatively minor impacts on the landscape and morphology, changes from sunny to overcast, cloudy weather, and snowy weather tended to be associated with major changes. BLIP captions showed that “overcast”, “cloudy”, and “snowy” weather were strongly associated with the word “mountains”, while “sunny” and “rainy” weather were associated with “woods” or “forest”. 

Seasonal descriptors can be challenging to evaluate since there is a strong local impact on seasonality. However, the generated images for the seasonal descriptors suggest a strong cold and/or temperate climate bias when using seasonal descriptors with Stable Diffusion. For the prompt “winter” 100\% of the images generated included snow, while 100\% of the images generated for the prompt “autumn” and the majority of images generated for “fall” included brightly colored foliage. The term “summer” had a much higher clip score (0.92) when compared with the other seasons, suggesting summer as the default season for river images. Similar to behaviour seen with the model including iconic national symbols for country name descriptors (Figure \ref{fig-3}), images generated for the prompts “spring” and “summer” often had flowering plants included in the image. This was also evident from the BLIP captions, with results for “spring” and “summer” both containing the word “flower”. The descriptor “fall” had a high failure rate relative to other seasonal descriptors, with many images of lakes or buildings rather than rivers, however, this behaviour was not seen with use of the synonym “autumn”.

\subsection{Image Reconstruction}
\begin{figure}
  \centering
  \includegraphics[width=0.8\textwidth]{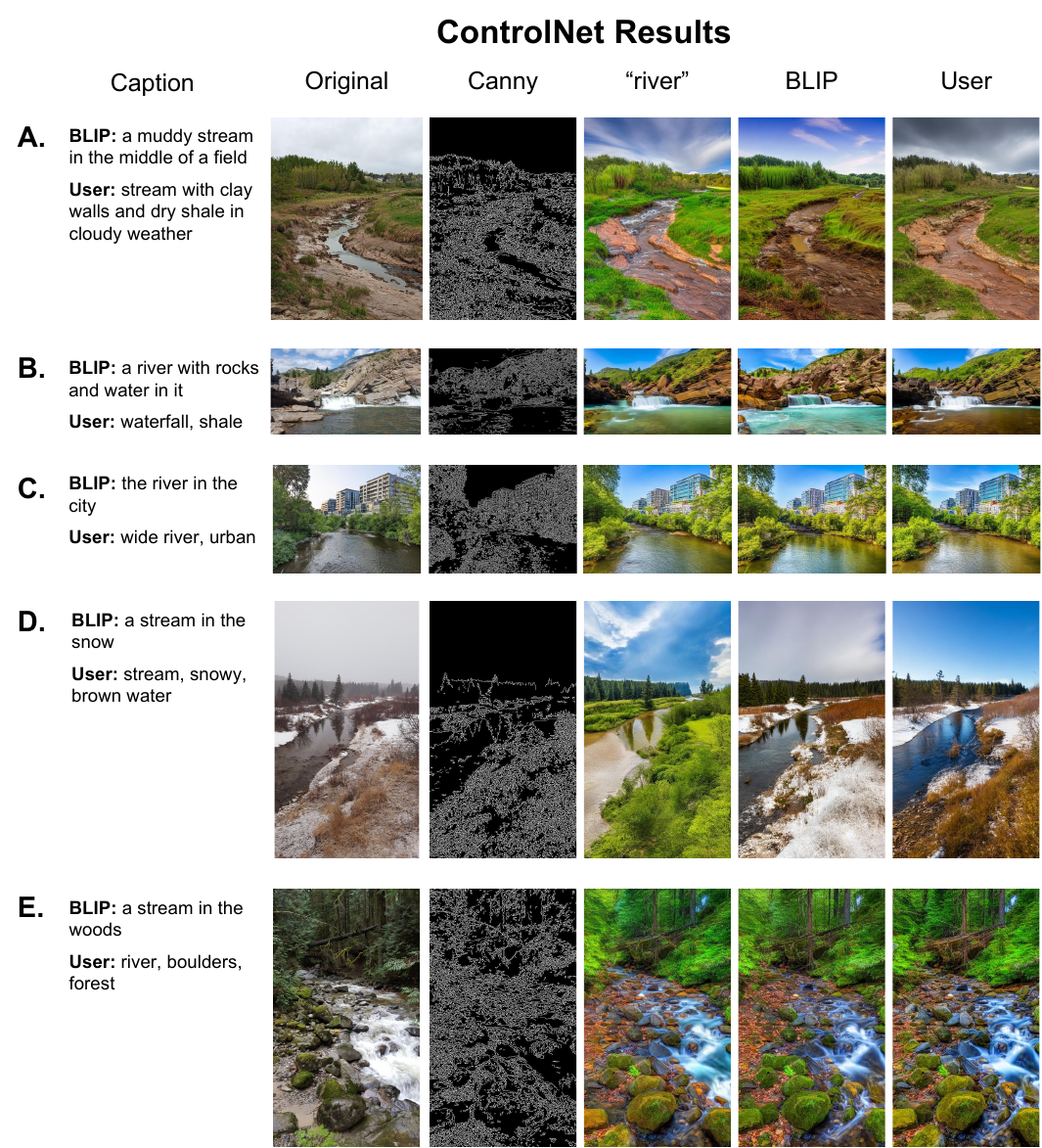} 
  \caption{Examples of the impact of prompt variation on image reconstruction in Stable Diffusion (v1.5) with ControlNet canny edge conditioning. Columns from left to right: captions, original images, canny edges, images generated for a default prompt “river”, images generated using the BLIP caption for the original image as a prompt, images generated using a custom user-defined prompt. Cherry-picked results from five images per prompt are shown.}
  \label{fig-6} 
\end{figure}

 Selected results from the image reconstruction task are shown in Figure \ref{fig-6}. Through the addition of Canny edge conditioning, it was generally possible to reproduce key morphological features of the input images including landscape, channel dimensions, channel gradient, and geology. Many of these features appear to be captured in the Canny edges (second column, Figure \ref{fig-6}), and in most cases, there was little difference between the prompt method used. However, for environmental features such as water color, weather, and season which are less likely to be observed in the Canny edges, a qualitative review of results suggests that the use of BLIP captions and user prompts are both able to produce much better results than the default prompt of “river”. The results in Figure \ref{fig-6} show cherry-picked results, where the best image from each batch of five generated images was selected. However, if all resulting images are considered, the BLIP captions tend to produce the most consistent results when compared with the other two prompting methods. 

\begin{figure}
  \centering
  \includegraphics[width=0.8\textwidth]{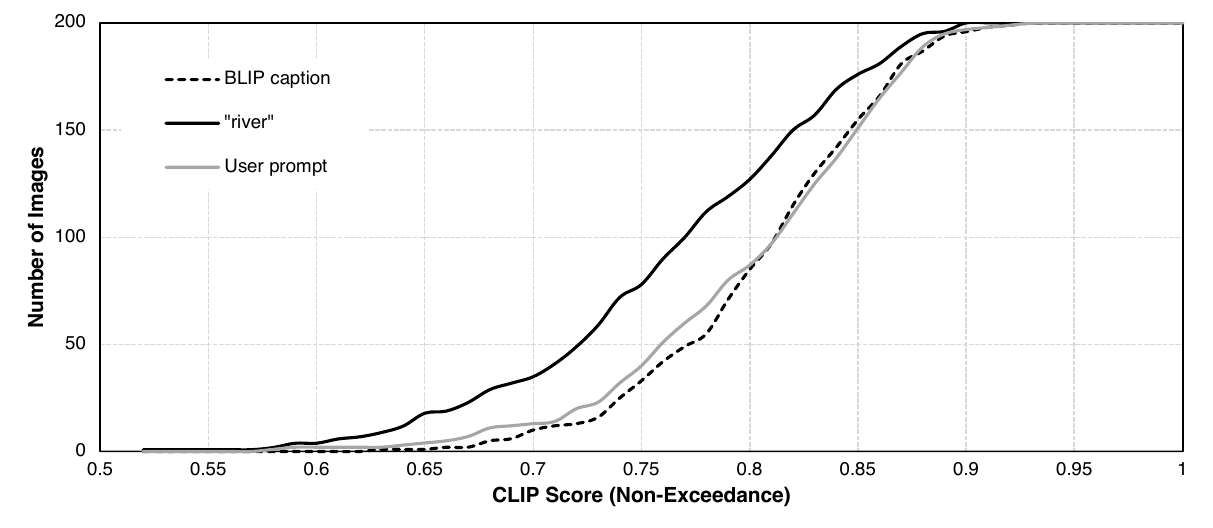} 
  \caption{CLIP Score non-exceedance plot for image reconstructions using Stable Diffusion (v1.5) with ControlNet Canny edge conditioning. Both BLIP captions and custom user-defined captions produced similar results which outperformed the default prompt of “river”. }
  \label{fig-7} 
\end{figure}

A quantitative comparison of the CLIP scores between the original images and generated images (Figure \ref{fig-7}) shows similar performance between the use BLIP captions and custom user-generated prompts, and confirms the qualitative finding of the two methods both outperforming the default prompt of “river”.  

\begin{figure}
  \centering
  \includegraphics[width=0.8\textwidth]{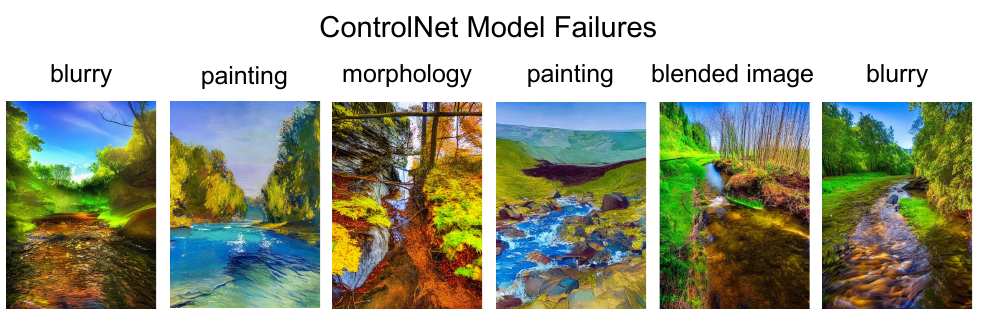} 
  \caption{Examples of common causes of failure for image reconstruction using ControlNet Canny edge conditioning with Stable Diffusion (v1.5).}
  \label{fig-8} 
\end{figure}

Many failures in image reconstruction were observed with the use of the ControlNet Canny edge model (Figure \ref{fig-8}), including images being generated in a painting style (despite this inclusion of this term as a negative prompt), the presence of blurry or unrealistic landscape and morphology features in images, or scenarios where two images appeared to be blended together. The use of BLIP captions as prompts resulted in the lowest failure rate and notably did not result in a single painting-style image being generated. The default prompt of “river” resulted in the highest failure rate, with user prompts performing between the two. 

\section{Discussion}
Key findings from the review of training data representation are consistent with the findings of previous studies conducted for non-river imagery \citep{Shankar2017-yf}, showing over-representation of western (i.e. North American, European) data. Scenic locations and popular tourist destinations were also over-represented, consistent with opportunistic and aesthetic biases identified in other environmental studies (e.g. \citep{Millar2019-bd}). Morphologically and environmentally, we identified key biases in the training data that had not been previously identified. For example, overrepresentation of waterfalls, which are often perceived as scenic, was noted, with the category containing more training images than all of the other evaluated gradient categories combined. 

In the image generation process, many of the impacts of overly aesthetic and opportunistic training data are evident in the generated images. For example, the changing of the larch (yellow foliage in Figure \ref{fig-2}B, Canada) is a short-lived and location-specific, but highly photographed event. Similarly, we hypothesize that the model’s difficulty in producing images of rivers for the prompt “rainy” is due to opportunistic data collection, with users avoiding taking pictures of rivers during precipitation events. 

Despite the general success of the Stable Diffusion model in generating river images, the current applications of this technology are somewhat limited. Several issues documented in the present study include the presence of road-style signs in images containing text similar to the prompts, watermarks in images, generation of not-safe-for-work images, oversaturation of output images, and generation of images that do not match the prompt description. Qualitative findings also suggest that the seed used in the image generation process is important, with certain seeds consistently producing images that better matched the prompts. Due to inconsistent outputs, we suggest that additional post-processing or manual inspection of images be used for sensitive applications. 

Fluvial geomorphologists, and more generally earth scientists who plan to use generative modelling to produce imagery should make themselves aware of differences in representation in the training data for the model, as well as familiarizing themselves with biases that exist in model outputs, such as preferences for mountain and forest landscapes, or strong associations with snow during winter. Interacting with these models can also be challenging, as the language used in the captions to train the models may not match the scientific jargon used by professionals. When generating images, users should also be aware of their own cognitive biases specific to the study area, such as historical preferences for specific landscapes or river morphologies \citep{Le_Lay2013-ui, Wilson2020-zk}. 

The use of additional conditional controls on image generation through tools such as ControlNet can help to mitigate some challenges associated with generative modelling. However, the use cases for this technology are somewhat more limited due to the requirement of additional inputs such as an edge or depth map. Although the exact process for generating edge-image-caption pairs during training of the Canny edge version of ControlNet is not detailed by the authors \citep{Zhang2023-kb}, several other versions of the ControlNet model released by the authors used BLIP to perform caption generation during the training process. Therefore, it is not surprising that the use of BLIP captions was very successful for image reconstruction, and generally able to match the performance of user inputs. Our findings suggest that auto-captioning of input images of rivers may be suitable in some cases. There is also good potential for users of the models to train their prompt engineering skills by learning from the BLIP captions to improve future image generation. 

However, images generated by ControlNet also suffered from several challenges, such as the frequent generation of painting style images despite the use of negative prompts to prevent this type of imagery. Qualitatively, the results from this study suggest that, on average, river images generated using ControlNet may be less realistic than images generated using the Stable Diffusion model without additional conditional controls. In the current study we used the default threshold for Canny edge detection, relatively short prompts, and consistent negative prompts across all images. We hypothesize that through additional prompt variation, including weighting of specific terms, and adjustment of Canny thresholds it may be possible to mitigate many of these challenges.  

Recent research has suggested that synthetic imagery from TTI models has now reached a point where it can be used for pre-training or transfer learning in few-shot-learning classification problems \citep{He2022-uu}. Our results showed that synthetic images generated by Stable Diffusion can reproduce the characteristics of many environmental and morphological descriptors commonly used in fluvial geomorphology, suggesting such use cases may be possible. However, due to strong biases in the results generated by the models, further testing would be required to understand how models trained on this synthetic data would perform on edge cases, where there is a known under-representation in original training data (e.g. African river during rainy weather). Additionally, improvements in the skill of semantic image-to-text generation models such as BLIP \citep{Li2022-rr} present an alternative to traditional ML-based classification techniques, allowing users to ask simple questions about images (e.g. “is it summer or winter?”). 

Although TTI models are continuing to increase in popularity, specific use cases for the field of fluvial geomorphology are not clear. Due to the data-scarce nature of the field, there is strong potential to use TTI models to create synthetic datasets for purposes such as pre-training machine learning models, for teaching and demonstration purposes. When Canny edge conditioning is added, other applications are also possible, such as seasonal augmentation of existing images (e.g. changing a picture of a river from summer to winter), image upscaling, or colorization of black and white or grayscale images.  

Beyond the Canny edges used in the current study, pre-trained ControlNet models are already able to accept other conditioning inputs such as line drawings or depth maps \citep{Zhang2023-kb}. In future work, additional models could be trained to accept conditioning maps more relevant to the field of fluvial geomorphology, such as digital elevation models, land use maps, soil types, or vegetation. Such technology has immeasurable potential for applications in the field. For example, generating synthetic river images from survey data, assisting with natural channel design, or modifying river images to show impacts of climate change (e.g. changes in flow, morphology, or vegetation) could all be achieved using TTI models with conditioning. 

As the technology becomes more mature and applications of the TTI in the earth sciences continue to expand, we expect cases to become more apparent.  

\section{Conclusions and Recommendations}
In the present study we identified biases in training data representation consistent with previous studies, including a preference for western culture and aesthetic/opportunistic data collection. Beyond these biases, we also identified key differences in representation specific to the field of fluvial geomorphology. Based on these findings, we recommend that prior to TTI models being used for other image generation tasks in the earth sciences, a similar review be undertaken to identify domain-specific challenges surrounding representation and biases. 

In general, we found that images of rivers generated using Stable Diffusion (v1.5) showed strong biases for specific morphological and environmental descriptors (e.g. mountain and forest landscapes, sunny weather, summer season). Despite these biases, we found that prompt variation was an effective technique for generating river images for many specific morphological and environmental characteristics. However, due to inconsistent model results, we recommend that additional post-processing or manual inspection of images be used for sensitive applications. 

Additional techniques for incorporating conditional control such as the use of ControlNet offer good potential for providing additional control over generated images. The Canny edge method used in this study was generally more effective in capturing morphological than environmental features of images. Due to the additional requirement of an edge map for image generation, use cases for these models are likely to have limited applications such as seasonal or weather augmentation in images. However, other conditioning maps offer great potential for additional use cases in fluvial geomorphology. 

Due to the rapid rise in popularity of TTI models, and relative ease of use (i.e. many cases requiring only a text prompt) we anticipate that the use of this technology will expand rapidly in the earth sciences field. There is great potential for these technologies to help rapidly advance many fields of study. However, we suggest cautious use in sensitive applications, and advocate that domain-specific reviews of training data and image generation biases for each model be conducted, as there is high potential for the use of TTI models to perpetuate existing biases. 

\section*{Acknowledgments}
This research was funded by the Natural Sciences and Engineering Research Council of Canada (NSERC) through the Discovery and CGS D programs. The researchers would also like to thank our summer student Lillian Collis for help with generating manual captions for images.
\bibliographystyle{agsm}

\bibliography{StableRivers}

\end{document}